\documentclass{article}

\usepackage{arxiv}

\usepackage[utf8]{inputenc} 
\usepackage[T1]{fontenc}    
\usepackage{tcolorbox}
\usepackage{multirow}
\usepackage{hyperref}       
\usepackage{url}            
\usepackage{amsmath}
\usepackage{booktabs}       

\usepackage{amsfonts}       
\usepackage{graphicx}
\usepackage{tikz}
\usetikzlibrary{shapes.geometric, arrows.meta}
\usepackage{xcolor}
\usepackage{times}
\usepackage{latexsym}
\usepackage{subcaption}
\usepackage{tikz}
\usetikzlibrary{shapes.geometric, arrows.meta, positioning}
\usepackage{xcolor}
\usepackage{pifont}
\usepackage{nicefrac}       
\usepackage{microtype}      
\usepackage{cleveref}       
\usepackage{lipsum}         
\usepackage{graphicx}
\usepackage{doi}
\usepackage{booktabs}
\usepackage{verbatim}
\usepackage{siunitx}
\usepackage{subcaption} 
\usepackage{array}
\usepackage{longtable}
\usepackage{geometry}
\usepackage{microtype}      
\usepackage{lipsum}
\usepackage{graphicx}
\graphicspath{ {./images/} }
\usepackage{amsmath}
\usepackage{framed}
\usepackage{tcolorbox}
\usepackage{amsmath}
\usepackage{times}
\usepackage{latexsym}
\usepackage{multirow}
\usepackage{listings}
\usetikzlibrary{arrows}

\usepackage[T1]{fontenc}
\usepackage{graphicx}
\usepackage{hyperref} 
\usepackage{subcaption}
\usepackage{tikz}
\usepackage[utf8]{inputenc}
\tcbuselibrary{listings}

\usepackage{graphicx}
\usepackage{tikz}
\usepackage{xcolor}
\usepackage{times}
\usepackage{latexsym}
\usepackage{tikz}
\usepackage{xcolor}
\usepackage{pifont}
\usepackage[utf8]{inputenc}
\usepackage{amsmath}
\usepackage{amsmath}
\usepackage{framed}
\usepackage{tcolorbox}
\usepackage{amsmath}
\usepackage{times}
\usepackage{latexsym}
\usepackage{multirow}
\usepackage{listings}
\usepackage{geometry}
\usepackage[numbers]{natbib}

\title{Mathematical Reasoning for Unmanned Aerial Vehicles: A RAG-Based Approach for Complex Arithmetic Reasoning}

\newif\ifuniqueAffiliation
\uniqueAffiliationtrue

\ifuniqueAffiliation 
\author{ \href{https://orcid.org/0000-0001-5597-4724}{\includegraphics[scale=0.06]{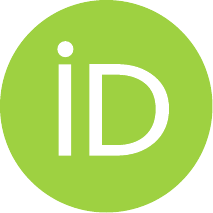}\hspace{1mm}Mehdi Azarafza}\\
	Department of Computer Science\\
	Hamm-Lippstadt University of Applied Sciences\\
	Lippstadt, Germany \\
	\texttt{mehdi.azarafza@hshl.de} \\
	\And
	\href{https://orcid.org/0000-0002-9177-0312}{\includegraphics[scale=0.06]{orcid.pdf}\hspace{1mm}Mojtaba Nayyeri} \\
	Institute For Artificial Intelligence\\
	University of Stuttgart\\
	Stuttgart, Germany \\
	\texttt{mojtaba.nayyeri@ipvs.uni-stuttgart.de} \\
	 \And
\href{https://orcid.org/0000-0002-2797-3828}{\includegraphics[scale=0.06]{orcid.pdf}\hspace{1mm}Faezeh Pasandideh} \\
Department of Computer Science\\
	Hamm-Lippstadt University of Applied Sciences\\
	Lippstadt, Germany \\
\texttt{faezeh.pasandideh@hshl.de}\\
\And
\href{https://orcid.org/0000-0002-0780-4154}{\includegraphics[scale=0.06]{orcid.pdf}\hspace{1mm}Steffen Staab} \\
Institute for Artificial Intelligence \\
University of Stuttgart, Germany \\
Stuttgart, Germany\\
\texttt{Steffen.Staab@ipvs.uni-stuttgart.de}\\
\AND
\href{https://orcid.org/0000-0002-9181-1110}{\includegraphics[scale=0.06]{orcid.pdf}\hspace{1mm}Achim Rettberg } \\
Department of Computer Science\\
	Hamm-Lippstadt University of Applied Sciences\\
	Lippstadt, Germany \\
\texttt{achim.rettberg@hshl.de}
\AND
}
\else
\usepackage{authblk}

\newbox{\orcid}\sbox{\orcid}{\includegraphics[scale=0.06]{orcid.pdf}} 
\author[1]{%
	\href{https://orcid.org/0000-0000-0000-0000}{\usebox{\orcid}\hspace{1mm}Mehdi  S.~Hippocampus\thanks{\texttt{hippo@cs.cranberry-lemon.edu}}}%
}
\author[1,2]{%
	\href{https://orcid.org/0000-0000-0000-0000}{\usebox{\orcid}\hspace{1mm}Elias D.~Striatum\thanks{\texttt{stariate@ee.mount-sheikh.edu}}}%
}
\affil[1]{Department of Computer Science, Cranberry-Lemon University, Pittsburgh, PA 15213}
\affil[2]{Department of Electrical Engineering, Mount-Sheikh University, Santa Narimana, Levand}
\fi

\renewcommand{\shorttitle}{RAG-UAV (Azarafza et al.)}

\hypersetup{
pdftitle={RAG-UAV},
pdfsubject={q-bio.NC, q-bio.QM},
pdfauthor={David S.~Hippocampus, Elias D.~Striatum},
pdfkeywords={Large Language Model, Retrieval-Augmented Generation, Unmanned Aerial Vehicle, UAV-Math-Bench},
}

\begin{document}
\maketitle

\begin{abstract}
Autonomous UAV operation necessitates reliable mathematical reasoning for tasks such as trajectory planning and power management.
While traditional flight control relies on hardcoded equations, recent Large Language Models (LLMs) offer potential for more flexible problem-solving but struggle with reliably selecting and applying
correct mathematical formulations and executing precise multi-step arithmetic. We propose RAG-UAV, a retrieval-augmented generation framework designed to improve the mathematical reasoning of several LLMs
(including GPT o1/Turbo, Llama-3.2/3.3, Mistral, and DeepSeek R1) in UAV-specific contexts by providing access to relevant domain literature. To conduct an initial assessment,
we introduce the UAV-Math-Bench, a 20-question problem set of UAV-centric mathematical problems across four difficulty levels.
Our experiments demonstrate that incorporating retrieval substantially increases exact answer accuracy (achieving up to 75\% with o1),
reduces instances of incorrect formulation selection (from 25\% without RAG to 5\% with RAG), and decreases numerical errors, reducing Mean Squared Error (MSE) by orders of magnitude for the best-performing models.
This pilot study indicates that RAG can enable general-purpose LLMs to function as more reliable tools for engineering analysis, although direct real-time flight control requires further investigation and validation on a
larger scale. All benchmark data, questions, and answers are publicly \footnote{https://github.com/Mehdiazarafza/UAV-RAG} available. 
\end{abstract}


\section{Introduction}

Unmanned Aerial Vehicles (UAVs) are aircraft that operate without an onboard pilot, enabling a wide range of applications from surveillance and mapping to delivery and emergency response. Autonomous UAV missions routinely demand multi-step mathematical reasoning—for instance, balancing energy budgets while re-planning swarms in variable wind. Existing flight-control code handles fixed equations, but
engineers
still spend time locating the right aerodynamic, communication-channel, or power-consumption formulas and checking the arithmetic.
This process demands a solid understanding of physics, systems engineering, and domain-specific formulas, and solid mathematical reasoning.

Large Language Models (LLMs), pretrained on vast corpora including physics texts and UAV design literature, offer a powerful new tool for assisting engineers. By leveraging their embedded knowledge, LLMs can help identify relevant equations, suggest performance evaluation strategies, validate reasoning steps, and even automate portions of UAV planning and analysis.
This positions LLMs as promising co-pilots in the engineering design loop, augmenting human expertise and accelerating development cycles.
While LLMs show promise for interpreting engineering queries and suggesting relevant design principles, they also exhibit important limitations. Their responses are constrained by the static nature of their pretraining data and may lack access to up-to-date, domain-specific documentation, standards, or experimental datasets. Moreover, LLMs may "hallucinate"—generating plausible but incorrect equations or assumptions, especially when faced with technical edge cases or incomplete prompts.
Retrieval-Augmented Generation (RAG) offers a practical solution by grounding LLM outputs in authoritative external sources. In this hybrid approach, relevant documents, retrieved at query time and used to condition the model’s response. This improves factual accuracy, enhances trustworthiness, and enables more context-aware assistance for engineers working on UAV design and operations.

\begin{figure*}[t]
    \centering
    \begin{tikzpicture}[node distance=1cm, every node/.style={scale=0.6}] 
   \node[draw, rounded corners, minimum width=5cm, minimum height=3cm, align=center] (stdinput1) {
    \textbf{o1 without RAG: } \\
  $P_{\text{profile}} \approx \frac{1}{8} \rho \sigma C_{D_0} \pi R^2 (\Omega R)^3 \quad \times \left[1 + k \left(\frac{v}{\Omega R}\right)^2\right]$
  
};
        \node[draw, rounded corners, minimum width=3cm, minimum height=1cm, align=center, below=of stdinput1] (stdout1) {
            \textbf{Model Output} \\
            A: The answer is Insufficient input data. \textcolor{red}{\ding{55}}
        };

        \node[draw, rounded corners, minimum width=5cm, minimum height=3cm, align=center, right=2cm of stdinput1] (cotinput1) {
            \textbf{o1 + RAG} \\
            $P_{\text{blade}}(V) = P_0 \left(1 + \frac{3V^2}{\Omega^2 R^2}\right).$
            
        };
        \node[draw, rounded corners, minimum width=3cm, minimum height=1cm, align=center, below=of cotinput1] (cotoutput1) {
            \textbf{Model Output} \\
            A: The answer is approximately  1.79. \textcolor{green}{\ding{51}}.
        };

        \draw[->] (stdinput1) -- (stdout1);
        \draw[->] (cotinput1) -- (cotoutput1);

    \end{tikzpicture}
    \caption{Comparison of Standard o1 and RAG with Prompting for Arithmetic Reasoning in UAV Applications}
    \label{fig:with_withoutRAG}
\end{figure*}
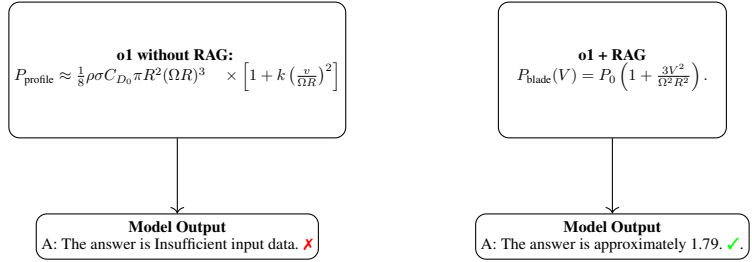

\subsection{Motivation Example}
Consider the task of computing the blade power of a UAV in forward flight, given specific parameters. As illustrated in Figure \ref{fig:with_withoutRAG}, a prominent LLM without access to external knowledge fails to identify the correct physical formulation, deeming the provided data "insufficient" (Appendix \ref{app:answerq1-2}). In contrast, when augmented with relevant domain literature via Retrieval-Augmented Generation (RAG), the same model successfully retrieves the appropriate equation, follows detailed reasoning steps, and computes an accurate result (Appendix \ref{app:correct1-2}). 
This example highlights a key challenge: LLMs may possess implicit knowledge but struggle with reliable access and application of domain-specific mathematical models, motivating our investigation into RAG as a solution.
Figure \ref{fig:with_withoutRAG} illustrates the results generated by the o1 model for the query:\\
"
Consider the following information:
\begin{itemize}
   \item The constant value $P_0$: 1.5
    \item The forward speed $v$: 110 km/h
    \item The blade angular velocity $\Omega$: 300 radians/second
    \item The rotor radius $R$: 0.4 m
    
\end{itemize}
Based on this information,  compute the Blade power of UAV in forward flight." \\

\subsection{Contribution}
In this work, we investigate how RAG can enhance the mathematical reasoning capabilities of LLMs for domain-specific UAV applications. We equip several LLMs with a vector-based retrieval mechanism over a corpus of UAV literature. Our contributions include:
\begin{itemize}
   \item RAG-UAV Framework: We propose and implement a RAG framework tailored for improving mathematical and physical reasoning in UAV scenarios.
   \item UAV-Math-Bench: We construct a categorized problem set of 20 UAV-centric mathematical reasoning problems across four difficulty levels to facilitate initial evaluation.
   \item Empirical Analysis: We conduct experiments comparing RAG-augmented and non-RAG LLM performance (ChatGPT-Turbo), quantifying improvements in answer accuracy, formulation selection, and numerical error across different models and problem complexities.
    \item Analysis of RAG Benefits: We demonstrate that RAG leads to substantial performance gains, particularly on more complex, multi-step problems, validating its potential for enhancing LLM-based engineering analysis in this domain. 
    
\end{itemize}

\section{Related Work}
We review the relevant literature in two key areas related to this work: (a) retrieval-augmented generation for reasoning, and (b) LLM in UAVs.

 \subsection{Retrieval-Augmented Generation for Reasoning} 
 Retrieval-augmented generation (RAG) is a method that combines language model generation with conventional information retrieval to extend the knowledge and reasoning limitations of LLMs.
 RAG was shown to significantly boost performance on knowledge-intensive NLP tasks by allowing a seq2seq model to consult a non-parametric memory (a Wikipedia index) during generation \cite{lewis2020retrieval}. 
 By conditioning on retrieved text, the model produced more accurate and specific outputs than a purely parametric baseline and even surpassed specialized retrieve-and-read pipelines on open-domain question answering. 
 RAG's general recipe is to embed the query, use vector similarity search to find semantically relevant documents, and then inject those documents (or snippets) into the model context. 
 
 While early applications of RAG focused on factual QA and knowledge retrieval, recent studies have adapted the concept to improve logical and mathematical reasoning. Other works have proposed related ideas, such as retrieving analogous solved problems to guide the model (a form of case-based reasoning) or using knowledge graphs to provide constraints during reasoning.
 All of these efforts point to the value of external knowledge in complementing the reasoning process of LLM. Retrieval can offer helpful anchors in the context of arithmetic, such as reminding the model of a relevant formula, a well-known arithmetic rule, or an example solution that it can adapt.  
 Our approach leverages this insight by providing each arithmetic query with related reference content drawn from a prepared corpus (papers), effectively bridging pure linguistic reasoning with algorithmic assistance from examples.

 \subsection{LLM in UAVs} 
 The intersection of LLMs and UAVs has opened new avenues for high-level autonomous decision-making in aerial robotics. Researchers have begun to integrate large language models into UAV control loops for tasks that require understanding complex instructions and making context-aware decisions. For example, TypeFly demonstrated an end-to-end system where an LLM translates natural language commands into drone flight actions, effectively piloting a quadcopter via language \cite{wang2025gsce}. 
 Similarly, the REAL framework incorporated an LLM as part of a UAV’s planning and control system, using the model to reason about goals and constraints during navigation \cite{tagliabue2024real}. 
 These approaches treat the LLM as a high-level “brain” that can interpret mission objectives, perform reasoning (e.g., break down a task into subtasks), and interface with low-level APIs or controllers. In multi-agent scenarios or human-UAV interaction, LLMs can serve as cognitive agents that understand dialogues or coordinate strategies among UAV teammates. Initial case studies report that LLM-driven UAVs can handle novel, complex directives more flexibly than pre-programmed systems. However, a recurring observation is that the reliability of such systems is fundamentally limited by the LLM’s reasoning fidelity. UAV control demands precise, logically consistent decisions, and LLMs, if not properly constrained, may produce outputs that violate physical constraints or 
 Our work aligns with this goal by tackling one particular weakness – mathematical reasoning – which is vital for UAV decision-making (e.g., calculating distances, timing sequences, or resource allocations). By improving mathematical accuracy through RAG, we aim to reduce one source of error in LLM-driven UAV systems, thereby contributing to safer and more trustworthy autonomous operations.

\section{Methodology}

\subsection{RAG-UAV Framework}
\begin{figure}[!ht]
\centering
\includegraphics[width=.7\linewidth]{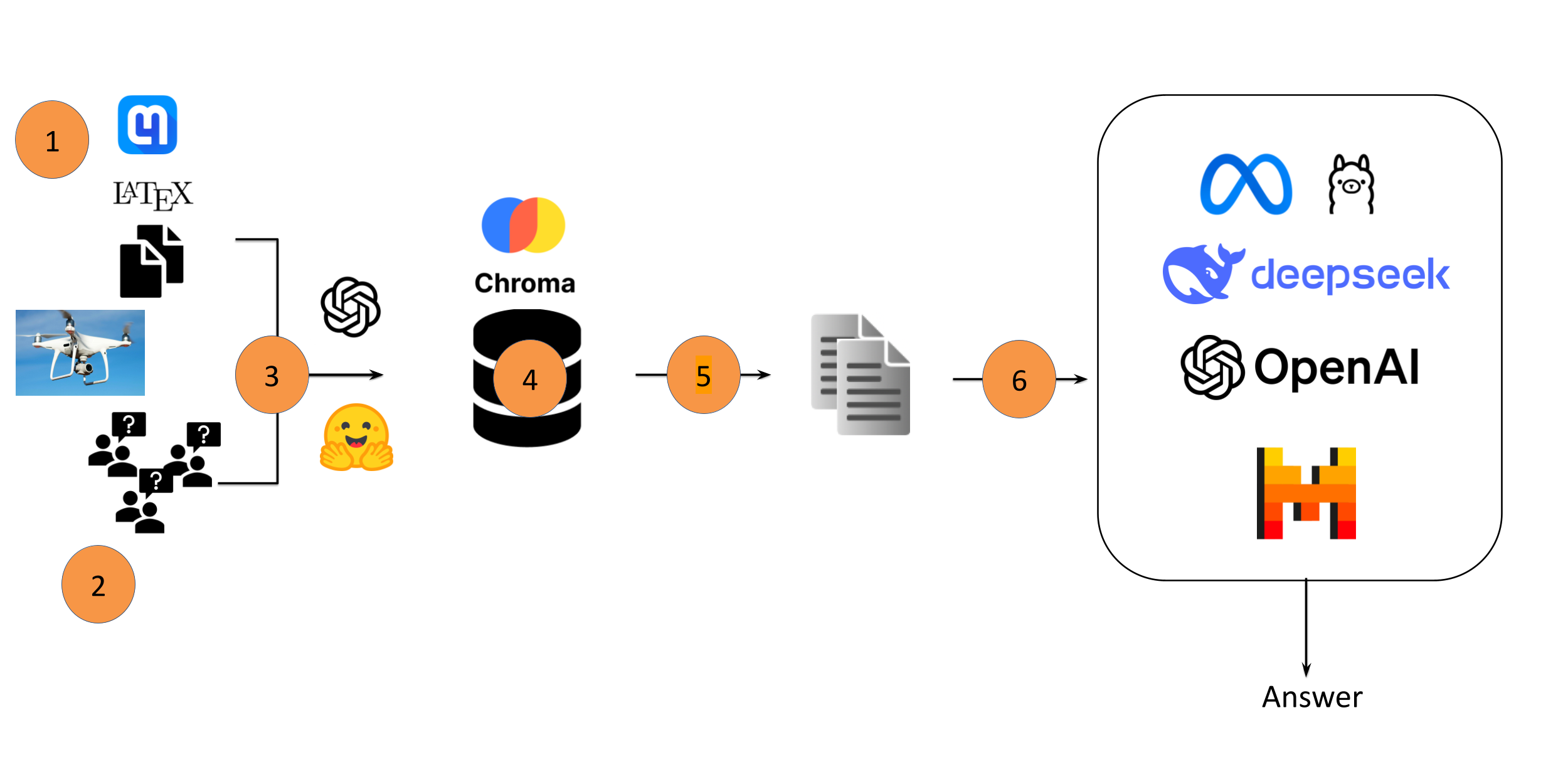} 
\caption{Workflow for RAG-UAV}
\label{fig:workload}
\end{figure}

We propose a RAG framework designed to augment LLMs with domain-specific mathematical knowledge. As illustrated in Figure \ref{fig:workload}, the framework operates in several stages:
(1) A user query is input, typically a mathematical problem related to UAVs. 
(2) The query is embedded into a vector representation using a chosen embedding model.
(3) This query embedding is used to perform a similarity search against a vector database containing embedded text segments from a corpus of UAV literature.
(4) The most relevant text segments are retrieved.
(5) The retrieved chunks are combined with the original user query into a structured prompt.
(6) This augmented prompt is fed to the LLM, which generates a response that includes the solution to the mathematical problem.
\subsection{Constructing UAVs' Vector Database}
Our knowledge base was constructed from a corpus of five research papers relevant to UAV mathematical modeling \cite{wang2022deployment,ryu2024energy,liu2022energy, yu2023backhaul,abubakar2023coverage}. 
Initially, our investigation involved processing PDFs directly fed to the models. However, we identified inaccuracies in the models' output that were subsequently attributed to errors during PDF parsing rather than inherent limitations in their core reasoning abilities. 
To accurately capture mathematical formulations, we processed the PDF documents using Mathpix\footnote{ https://mathpix.com/} to convert them into LaTeX format. This process aimed to preserve equations and symbols more accurately than direct PDF parsing. The resulting LaTeX output was segmented into page-level chunks, yielding a total of 162 segments for the five papers. These segments were then embedded using two different models: all-mpnet-base-v2 (768 dimensions)\cite{HF-mpnet} for open-source LLMs and OpenAI's text-embedding-ada-002 (1536 dimensions)\cite{Zilliz-Ada002} for OpenAI models. The resulting vector representations were stored in a Chroma vector database.

where  \cite{wang2022deployment}

For running open-source models, we used Ollama \cite{ollama_2025} and for OpenAI models, we used the API.

\subsection{LLM Prompting Strategy}
The prompt template used to query the LLMs incorporates both the original problem statement and the retrieved text segments. The template instructs the model to utilize the provided documents for identifying relevant formulations but allows the model to use its internal capabilities for performing the numerical calculations. A simplified structure of the prompt is: "Use only the provided documents: [Retrieved Documents]. Question: [Original Question]. This strategy was chosen to isolate the effect of retrieval on formulation selection while allowing the LLM to perform the numerical computation, leveraging its inherent arithmetic abilities. It also reflects a potential real-world use case where an LLM might be provided with domain knowledge but still rely on its internal calculation engine. The full prompt template is provided in Appendix \ref{appendix:prompt}.

\subsection{UAV-Math-Bench: Evaluation Problem Set}
To evaluate the complexity of computational tasks and analytical reasoning required in solving the provided problems, the questions are categorized into four distinct levels: Basic, Intermediate, Advanced, and Expert. 
The evaluation set serves to demonstrate the potential of the RAG-UAV framework and highlight performance differences between models and task complexity.
This classification considers the number of parameters involved, the mathematical complexity, and the depth of domain-specific knowledge required to obtain a solution. The development of UAVs is intrinsically linked to this increasing complexity. 
As UAV technology advances, the challenges associated with their deployment grow more intricate, requiring sophisticated problem-solving techniques, advanced mathematical models, and a deep understanding of engineering principles. The categorization of difficulty levels thus reflects the rising complexity of real-world UAV-related problems encountered by professionals today.

\subsubsection{Basic Level Questions}

Basic level questions require direct computation based on a given set of parameters, typically involving straightforward mathematical formulas. These problems involve fundamental arithmetic or algebraic manipulations without requiring extensive domain knowledge.

\begin{itemize}
    \item \textbf{Q1-1:} Computation of elevation angle speed based on coordinate positions.
    \item \textbf{Q1-2:} Calculation of blade power using given constants and basic aerodynamics equations.
    \item \textbf{Q1-3:} Computational latency determination based on CPU allocation and task parameters.
    \item \textbf{Q1-4:} Atmospheric attenuation factor computation using given wavelength and scattering distribution.
    \item \textbf{Q1-5:} Received power computation using path loss and channel gain formulas.
\end{itemize}

At this level, UAVs are typically used for simple tasks such as altitude measurement and power consumption estimation, often relying on basic physics and aerodynamics.

\subsubsection{Intermediate Level Questions}

Intermediate questions extend beyond direct computations and require additional steps, such as probability computations or logarithmic transformations. These problems involve a moderate level of engineering knowledge and mathematical formulation.

\begin{itemize}
    \item \textbf{Q2-1:} Probability computation of a line-of-sight (LoS) connection based on given probability constants.
    \item \textbf{Q2-2:} Induced power computation in forward UAV flight using additional aerodynamics principles.
    \item \textbf{Q2-3:} Data transmission rate calculation using Shannon’s theorem.
    \item \textbf{Q2-4:} Achievable data rate computation requiring logarithmic calculations.
    \item \textbf{Q2-5:} Power consumption due to communication, incorporating multiple power-related parameters.
\end{itemize}

At this intermediate level, UAVs become more complex in their operations, handling tasks such as communication protocols, energy management, and data transmission in various environmental conditions. 
\subsubsection{Advanced Level Questions}

Advanced questions involve multi-step problem-solving approaches that integrate multiple mathematical principles. These computations require a solid grasp of signal processing, aerodynamics, and wireless communication concepts.

\begin{itemize}
    \item \textbf{Q3-1:} Path loss calculation for a LoS connection incorporating frequency and speed of light.
    \item \textbf{Q3-2:} Parasite power computation based on aerodynamic drag and air density factors.
    \item \textbf{Q3-3:} Downlink transmission rate computation using transmission bandwidth and noise power.
    \item \textbf{Q3-4:} Bandwidth allocation to meet a specific data rate requirement.
    \item \textbf{Q3-5:} Power consumption due to hovering, integrating multiple aerodynamic and energy parameters.
\end{itemize}

In this phase, UAVs are employed for complex tasks such as networked communication between UAVs, long-duration flights, and advanced environmental monitoring, where power efficiency and reliable data transmission are critical.

\subsubsection{Expert Level Questions}

Expert-level questions demand an extensive understanding of multiple concepts and their integration into a single computational framework. These problems often require optimization, iterative solutions, or advanced probabilistic and statistical analyses.

\begin{itemize}
    \item \textbf{Q4-1:} Average path loss computation involving LoS and NLoS path loss models.
    \item \textbf{Q4-2:} Total UAV power consumption in forward flight, integrating multiple power components.
    \item \textbf{Q4-3:} Transmission delay computation based on bandwidth, power, and noise parameters.
    \item \textbf{Q4-4:} Utility ratio determination requiring multi-variable calculations and resource allocation.
    \item \textbf{Q4-5:} Comprehensive UAV power consumption computation incorporating aerodynamic, communication, and electrical parameters.
\end{itemize}

At the expert level, UAVs are operating in highly sophisticated environments, with tasks requiring optimal energy management, high-level coordination, and complex decision-making algorithms, often for missions such as autonomous fleet management or large-scale environmental surveying \cite{PASANDIDEH2025103767}.

Table \ref{tab:question_categorization} categorizes the questions by difficulty, with additional details on the number of variables and nested equations.

\begin{table}[t]
\caption{Categorization of questions by difficulty level, number of variables, and nested equations.}
\centering
\small
\begin{tabular}{llll}
\toprule
\textbf{Difficulty} & \textbf{Question} & \textbf{\# Vars} & \textbf{\# Nest. Eqns} \\
\midrule
\multirow{5}{*}{Basic} 
  & Q1-1 & 2 & 1 \\
  & Q1-2 & 3 & 1 \\
  & Q1-3 & 2 & 1 \\
  & Q1-4 & 2 & 1 \\
  & Q1-5 & 3 & 1 \\
\midrule
\multirow{5}{*}{Intermediate} 
  & Q2-1 & 2 & 2 \\
  & Q2-2 & 4 & 2 \\
  & Q2-3 & 3 & 1 \\
  & Q2-4 & 3 & 1 \\
  & Q2-5 & 4 & 2 \\
\midrule
\multirow{5}{*}{Advanced} 
  & Q3-1 & 4 & 2 \\
  & Q3-2 & 5 & 2 \\
  & Q3-3 & 4 & 2 \\
  & Q3-4 & 3 & 2 \\
  & Q3-5 & 5 & 3 \\
\midrule
\multirow{5}{*}{Expert} 
  & Q4-1 & 6 & 3 \\
  & Q4-2 & 6 & 3 \\
  & Q4-3 & 5 & 3 \\
  & Q4-4 & 6 & 4 \\
  & Q4-5 & 7 & 4 \\
\bottomrule
\end{tabular}
\label{tab:question_categorization}
\end{table}

\section{Evaluation and Results}
This section presents the experimental setup and results demonstrating the impact of RAG on the mathematical reasoning performance of various LLMs on the UAV-Math-Bench. We compare RAG-augmented models against ChatGPT non-RAG baselines and other models with RAG using three primary metrics: Exact Answer Accuracy, Formulation Selection Error Rate, and Mean Squared Error (MSE).
\subsection{Evaluation Metrics}

\textbf{Exact Answer Accuracy:} This metric measures the percentage of problems for which the model produces a final numerical answer that precisely matches the ground truth answer provided by human experts. Answers must match within a predefined tolerance for floating-point comparisons ( $10^{-4}$), where applicable.

\textbf{Formulation Selection Error Rate:} 
This metric quantifies the percentage of problems where the model either selects an incorrect mathematical formula or fails to identify any relevant formula from the provided context.
This is assessed through the expert manual inspection of the model's step-by-step reasoning process.

\textbf{Mean Squared Error (MSE):} We calculate the MSE between the model's final numerical answer and the ground truth answer across the benchmark. Due to potentially large differences between model outputs and ground truths, we report results using $\log_{10}(\text{MSE})$ for clearer visualization. A lower MSE indicates closer agreement with the reference answer.

\subsection{Comparison of models performance }
To ensure a comprehensive comparison, we employed the correct formulations and verified the answers in all LLM. The complete set of answers, the corresponding questions and the details of the model are available on the GitHub repository of the paper.
\begin{table}[t]
\centering
\small
\caption{Comparison of performance metrics across evaluated LLMs.}
\label{tab:model_comparison}
\begin{tabular}{lcccccccc}
\toprule
\textbf{Metric} & \textbf{Chat} & \textbf{o1} & \textbf{o1-p} & \textbf{Turbo} & \textbf{Mist} & \textbf{L-3B} & \textbf{R1} & \textbf{L-70B} \\
\midrule
Correct (\%) & 30.00 & 75.00 & 70.00 & 60.00 & 5.00 & 10.00 & 50.00 & 35.00 \\
Correct Form. (\%) & 75.00 & 95.00 & 85.00 & 75.00 & 70.00 & 60.00 & 90.00 & 65.00 \\
Miscalc. (\%) & 45.00 & 15.00 & 10.00 & 20.00 & 55.00 & 50.00 & 35.00 & 30.00 \\
MisC$^\dagger$ (\%) & 60.00 & 21.05 & 11.76 & 26.67 & 78.57 & 83.33 & 38.89 & 46.15 \\
\bottomrule
\end{tabular}
\vspace{0.5em}
\raggedright
\footnotesize
\\$^\dagger$MisC = Miscalculation among correctly formulated answers. \\
Chat = ChatGPT Turbo, L-3B = LLaMA3.2 3B, L-70B = LLaMA3.3 70B, R1 = DeepSeek R1, Mist = Mistral 7B. \\
Miscalc. = Miscalculation (out of total attempts). \\
\end{table}

Table \ref{tab:model_comparison} compares the performance of several LLMs based on key metrics. We evaluated both correct and wrong answers, and if a model did not answer or asked for more information, we counted that as zero. ChatGPT is the only model without RAG. Among the models, ol performed the best with correct answers 75\% and the lowest error in the calculations. OpenAI's o1 model consistently identified the correct formulation and computed values close to the reference answers, resulting in a considerably lower MSE compared to others. This model also provided optimized and well-structured answers, demonstrating high reasoning capabilities in complex questions. Notably, OpenAI's o1 model demonstrated a degree of semantic robustness, capable of interpreting physical constants such as the speed of light despite variations in user input, including misspellings and non-standard variable assignments. For instance, when presented with the statement defining 'G' as the "speed light", the model inferred the intended concept, correlating 'c' with the symbol for the speed of light.\\
Models like Mistral 7B and LLaMA 3.2 3B gave mostly wrong answers and made many mistakes. Mistral 7B demonstrated lower performance overall. DeepSeek 70B\cite{deepseek} and LLaMA 3.3 70B gave more accurate answers than smaller models, and DeepSeek made fewer mistakes overall. DeepSeek provides very detailed and precise answers (Appendix\ref{app:deepseekq1-1}) and is considered the best open-source LLM model in our evaluation. ChatGPT-4-Turbo without rang and Turbo with RAG were in the middle - they could answer some questions but often miscalculated. This shows that larger models and better training are important in answering technical questions such as those about RAG UAVs.\\
The o1-preview model demonstrated the most robust performance, exhibiting a low miscalculation rate of 11.76\% when correct formulations were achieved. In contrast, Mistral 7B and Llama 3.2 3b showed markedly higher miscalculation rates (78.57\% and 83.33\%, respectively), suggesting a significant challenge in maintaining computational accuracy for these models.

\begin{figure}[!ht]
\centering
\includegraphics[width=\linewidth]{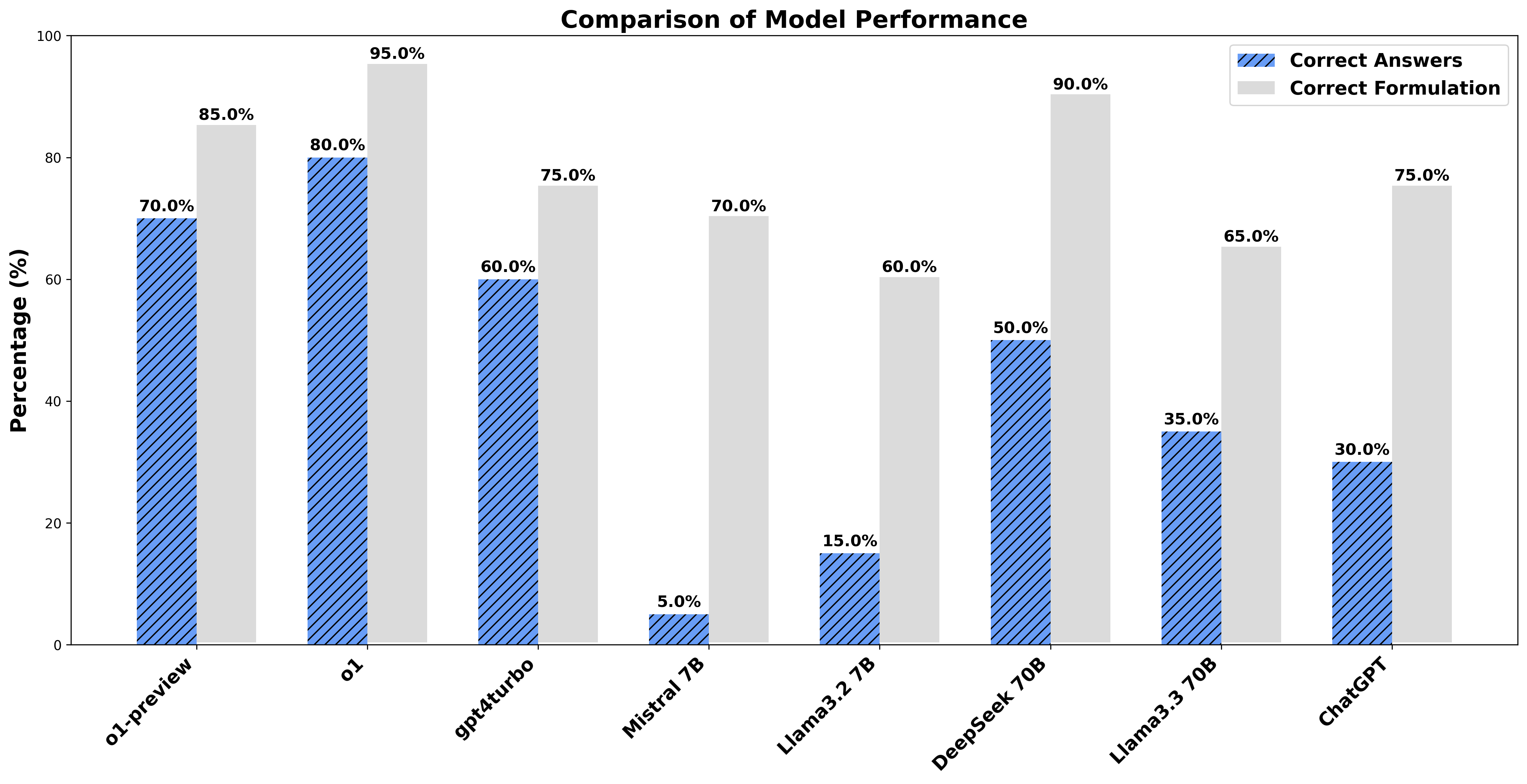} 
\caption{Compare state-of-the-art LLMs  }
\label{fig:compare}
\end{figure}
\begin{figure}[!ht]
\centering
\includegraphics[width=.7\linewidth]{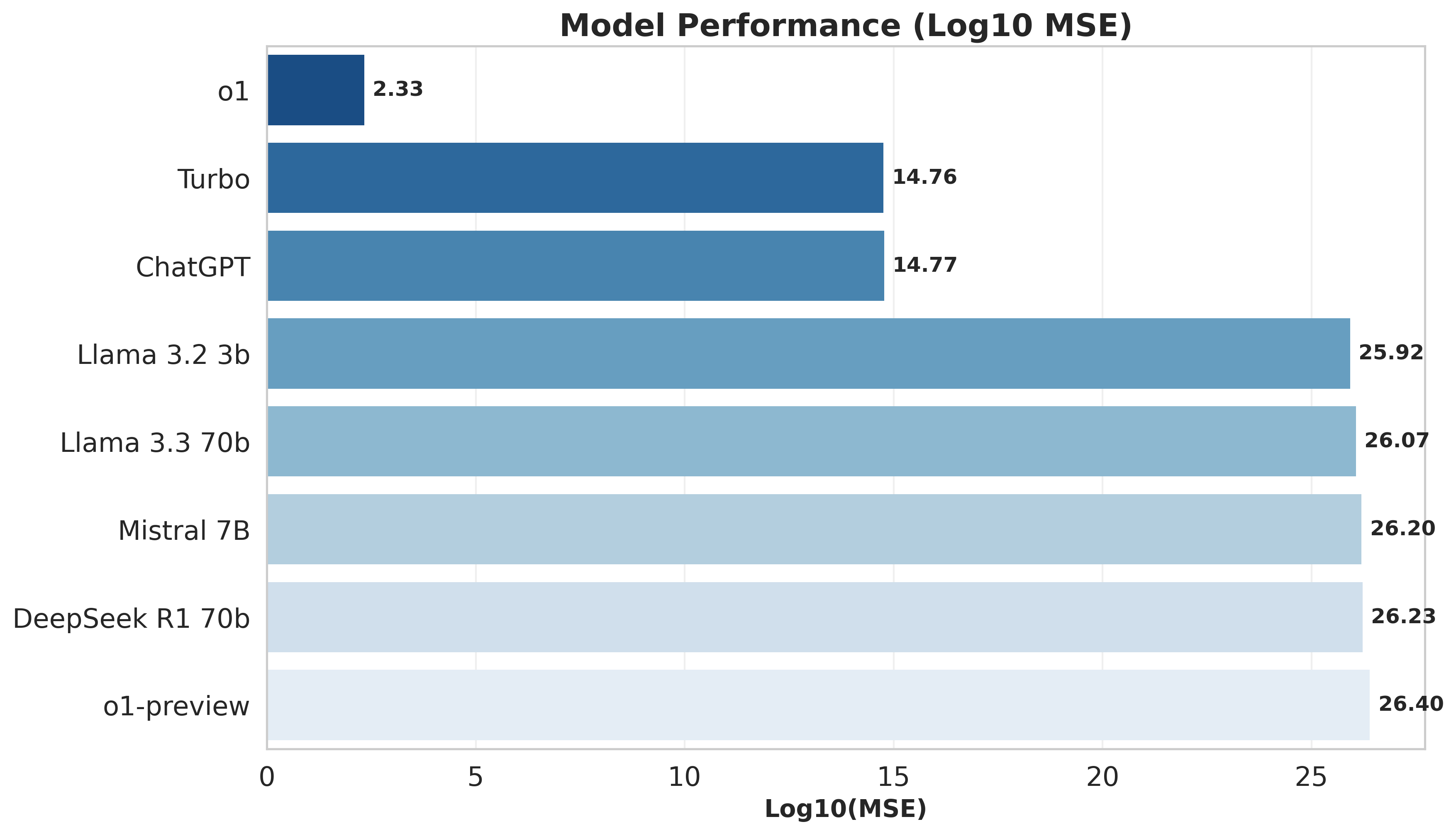} 
\caption{MSE Comparison (Log10(MSE)) }
\label{fig:mse}
\end{figure}

Figure \ref{fig:mse} compares the performance of several LLMs based on MSE. Since MSE values are extremely large due to the significant differences between human responses and model predictions, we used a log 10 scale to make the evaluation clearer. The results show that o1 outperforms all other models dramatically with an MSE of just  214.74. \\In particular, significant deviations can occur depending on the model's ability to select and apply the correct formulation. For example, in one benchmark question, the expert-provided ground truth is 64,207,051.66, while ChatGPT returned a significantly lower value of 70.2. Similar differences are observed in several models. However, the o1 model consistently identified the correct formulation and computed values close to the reference answers, resulting in a considerably lower MSE compared to others.
\subsection{Performance by Difficulty Level}
In this section we provide a comparative evaluation of eight large language models across four difficulty tiers, summarized in Table \ref{tab:diff_compare}. 

\begin{table}[t]
\centering
\small
\caption{Accuracy comparison across difficulty levels for evaluated models.}
\label{tab:diff_compare}
\begin{tabular}{lcccccccc}
\toprule
\textbf{Level} & \textbf{Chat} & \textbf{o1} & \textbf{o1-p} & \textbf{Turbo} & \textbf{M7B} & \textbf{L3-3B} & \textbf{R1} & \textbf{L3-70B} \\
\midrule
Basic     & 60\% & 80\% & 100\% & 100\% & 20\% & 0\%  & 60\% & 60\% \\
Intermediate & 0\%  & 60\% & 80\%  & 60\%  & 0\%  & 20\% & 40\% & 60\% \\
Advanced  & 40\% & 80\% & 60\%  & 60\%  & 0\%  & 0\%  & 80\% & 20\% \\
Expert    & 20\% & 80\% & 40\%  & 20\%  & 0\%  & 20\% & 20\% & 0\% \\
\bottomrule
\end{tabular}
\vspace{0.5em}

\raggedright
\footnotesize
Chat = ChatGPT, o1-p = o1-preview, Turbo = GPT-4 Turbo, M7B = Mistral 7B, \\
L3-3B = LLaMA 3.2 3B, R1 = DeepSeek R1, L3-70B = LLaMA 3.3 70B.
\end{table}
The o1 model outperforms all others, achieving 80\% accuracy even on Expert-level questions. While most models handle Basic questions well, performance drops notably for harder levels. Models like Mistral 7B and LLaMA 3.2 3B perform poorly overall, especially on Advanced and Expert tasks. ChatGPT and DeepSeek R1 show moderate accuracy, but only o1 maintains consistently high performance across all levels.
Interestingly, the O1 model achieved 80\% accuracy on Expert-level problems, a performance not mirrored at the Intermediate level, where two errors occurred. One specific instance involved a slight numerical deviation in the model's output (0.5535 vs. the correct 0.5), classified as incorrect under rigorous human expert assessment, even though the model itself indicated the result was 'approximately 0.50.' For another mistake from the model, there was a unit conversion error: we assumed $\theta$ is in radians and applied a conversion to degrees, but o1 already considered the angle in degrees. This led to an incorrect input to the exponential term and significantly altered the result. \\
While Expert problems involve more complex chains of reasoning and integration of multiple concepts, the model might excel at retrieving and structuring these high-level solutions. However, it may be less robust to handling refined details like precise numerical tolerances or implicit unit conventions that were critical for correctness in some Intermediate problems.

\subsection{Examination of Attention Patterns}
\begin{figure}[ht]
  \centering
  \begin{subfigure}[b]{0.48\linewidth}
    \includegraphics[width=\linewidth]{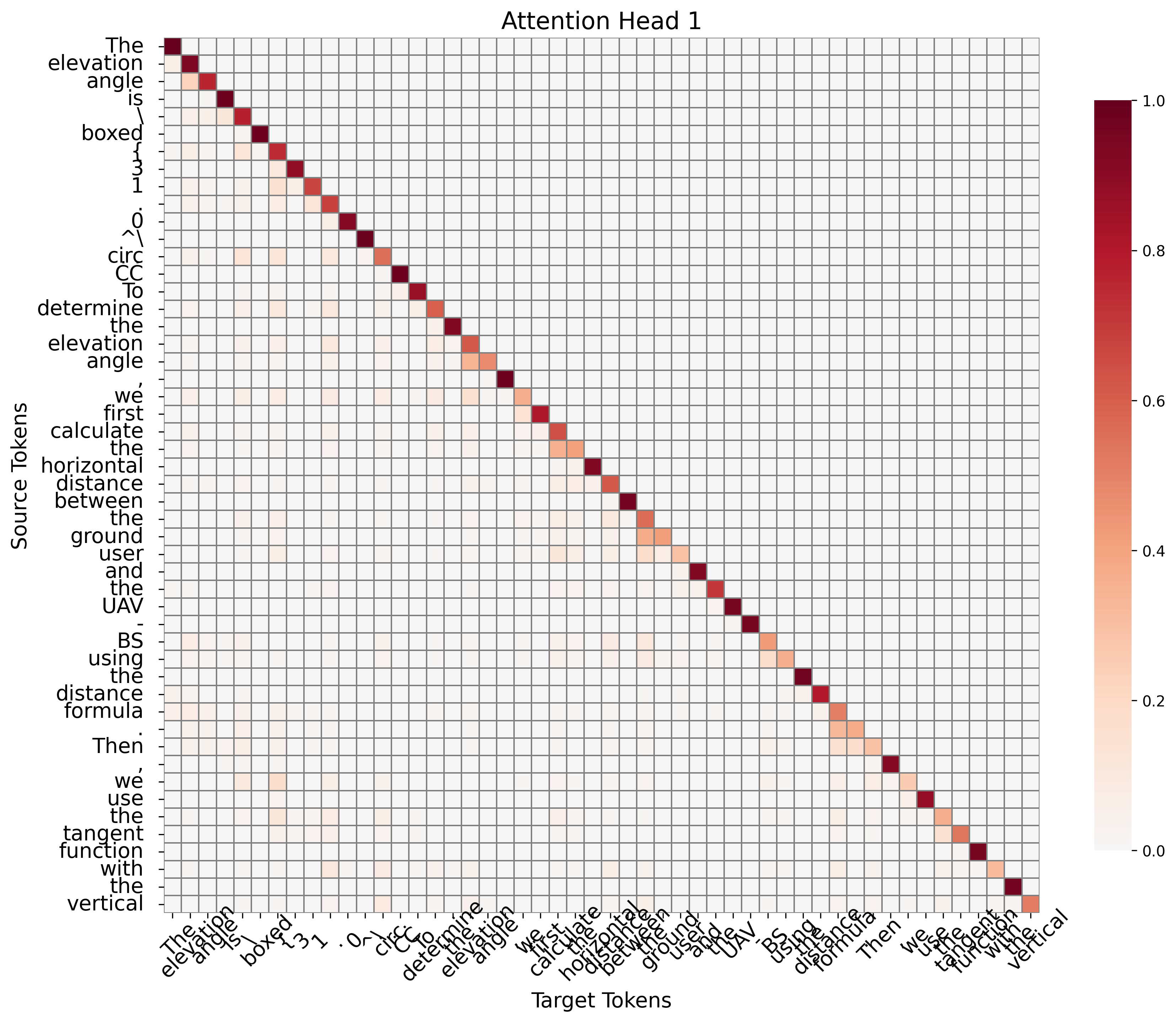}
    \caption*{(a) Attention Matrix: Head 1 of DeepSeek R1}
  \end{subfigure}
  \hfill
  \begin{subfigure}[b]{0.48\linewidth}
    \includegraphics[width=\linewidth]{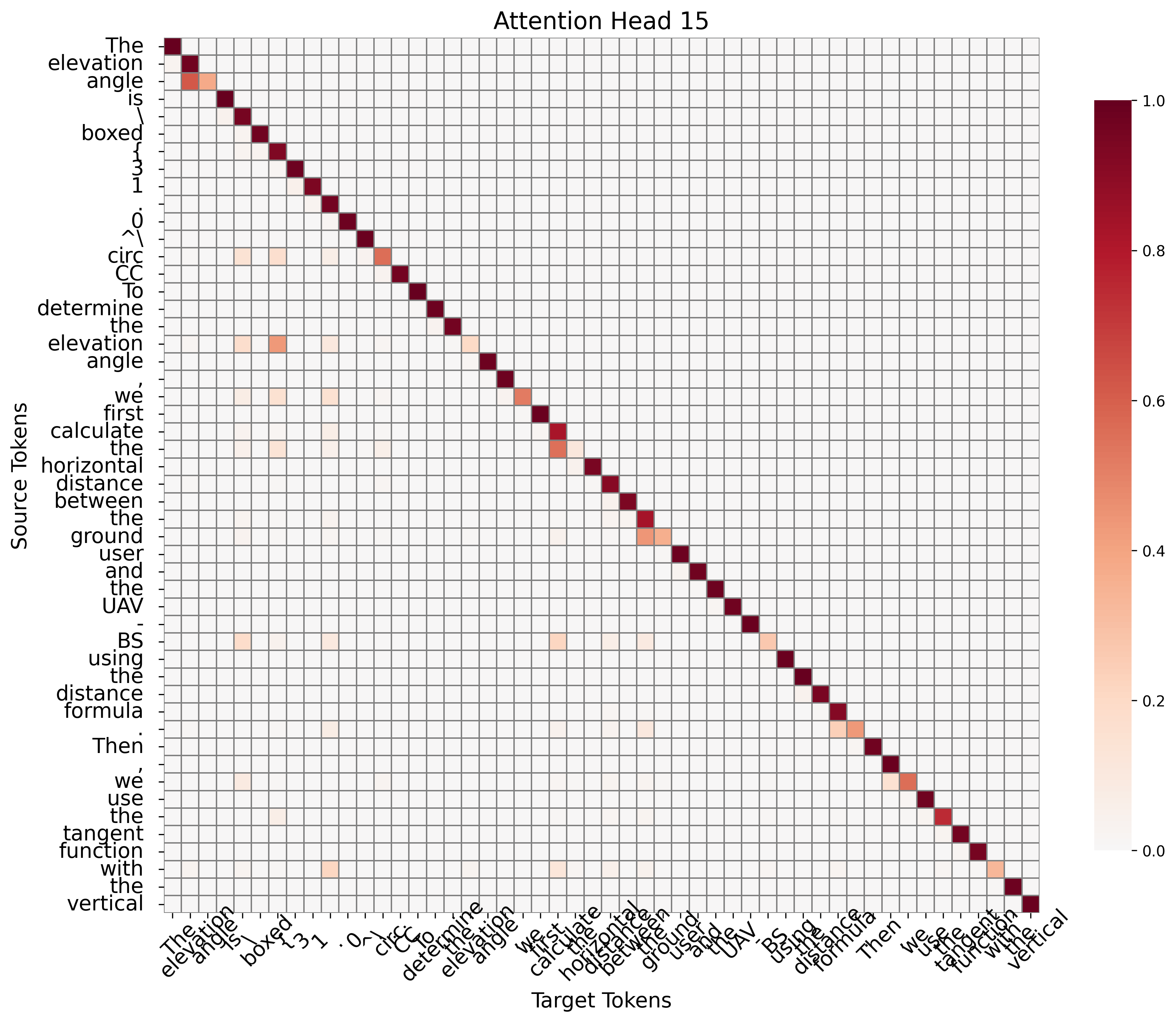}
    \caption*{(b) Attention Matrix: Head 15 of DeepSeek R1}
  \end{subfigure}
  \caption{Attention matrix visualization for different heads in DeepSeek R1 (Layer 1).}
  \label{fig:head1}
\end{figure}

Understanding the internal processes of LLMs, particularly in complex tasks like mathematical reasoning, remains an active research area \cite{gao2023retrieval}. To gain preliminary insights into how the Deep Seek R1 model processes input and potentially leverages information, we visualized the attention patterns of its first layer for a selected question and its generated answer. \\ 
Figure \ref{fig:head1} shows attention patterns for two different heads for first layer, revealing similar diagonal and localized attention structures within the generated sequence.

\begin{figure}[!ht]
\centering
\includegraphics[width=.7\linewidth]{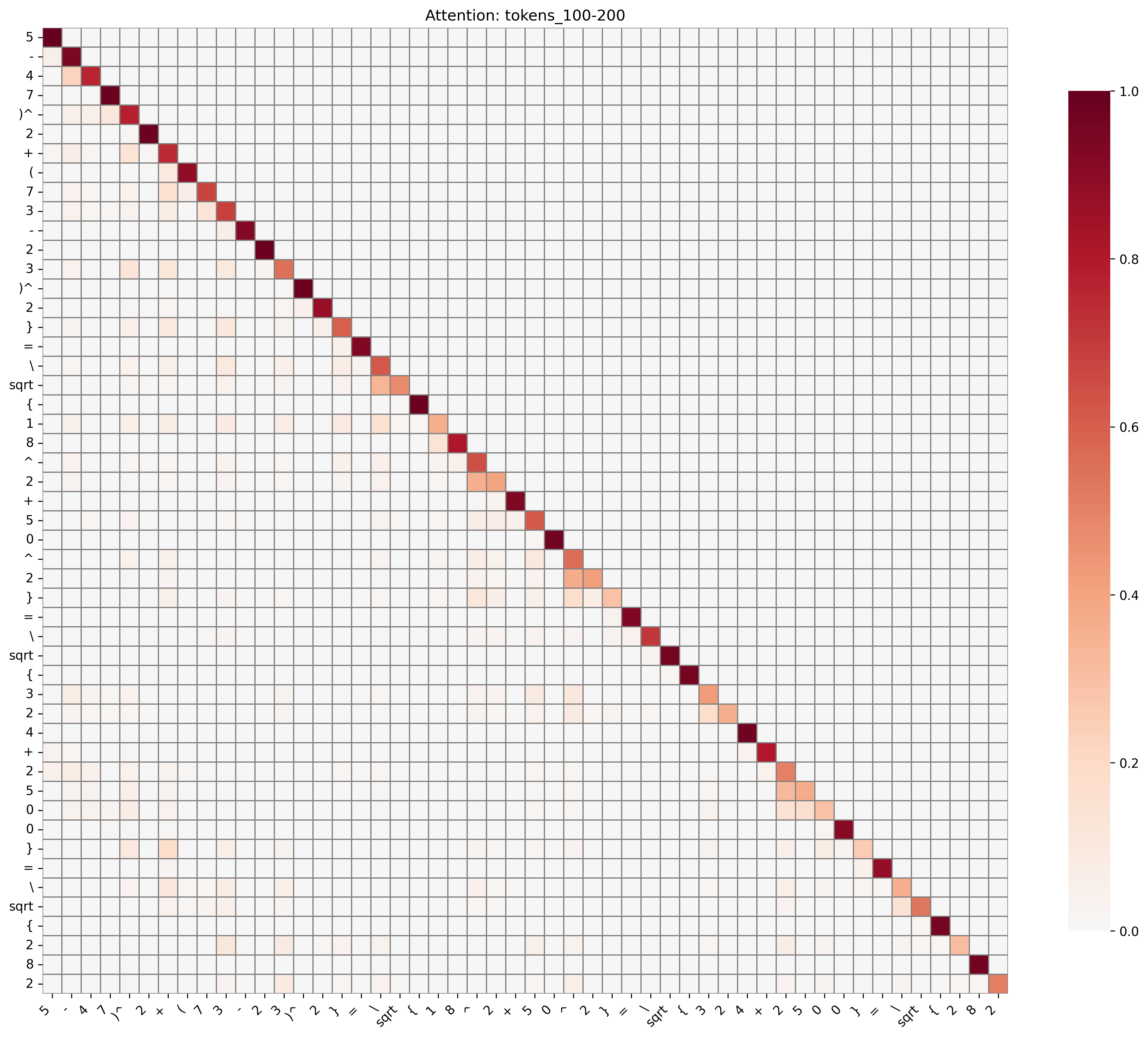} 
\caption{Distance Calculation Attention Weights}
\label{fig:token200}
\end{figure}

\begin{figure}[!ht]
\centering
\includegraphics[width=.7\linewidth]{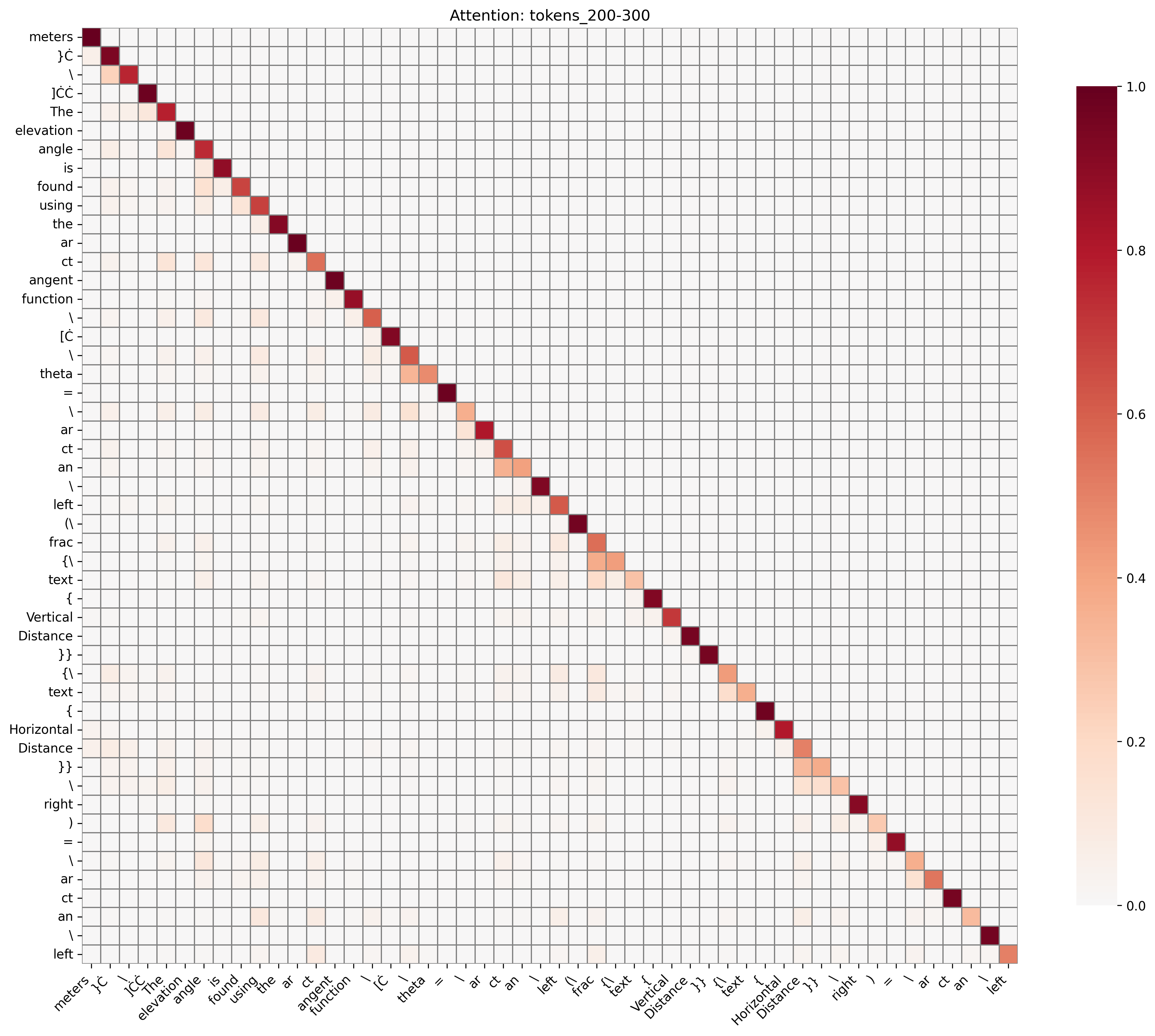} 
\caption{Equation Finding Attention Weights}
\label{fig:token300}
\end{figure}

Figure \ref{fig:token200} highlights attention weights during the calculation of the horizontal distance, showing the model attending to relevant numerical tokens from the input.
 \begin{equation*}
\text{Horizontal distance} = \sqrt{(65 - 47)^2 + (73 - 23)^2}
\end{equation*}
Figure \ref{fig:token300} illustrates the tokens involved in calculating theta, illustrating that the model also considers the previously computed token "horizontal distance" in its calculations. 


\clearpage
\section{Conclusion}
In this work, we demonstrated that retrieval-augmented generation (RAG) can substantially improve the mathematical reasoning capabilities of Large Language Models (LLMs), particularly within the context of domain-specific UAV applications requiring access to physical models and accurate multi-step computations. By providing LLMs with relevant external knowledge via RAG, we observed a significant increase in their ability to correctly select formulations and compute accurate numerical results on a set of UAV-centric mathematical problems.

Our experiments showed consistent gains in accuracy and substantial reductions in numerical error across several LLMs when augmented with RAG. These improvements were most pronounced on problems involving multiple steps and nested operations, suggesting that RAG effectively mitigates common failure modes of standalone LLMs on complex mathematical tasks. While this study utilizes a limited problem set, the findings indicate that RAG holds significant potential for enabling LLMs to function as more reliable tools for engineering analysis in technical domains like UAV development, by providing them with the necessary domain knowledge to perform complex calculations more accurately. \\\\

\section*{Limitations}
This work demonstrates that Retrieval-Augmented Generation (RAG) substantially enhances Large Language Models' (LLMs) mathematical reasoning, improving accuracy and reducing errors in UAV-specific multi-step computations by providing relevant external knowledge. However, this study's findings are limited by its small benchmark size and further practical deployment requires rigorous evaluation of the RAG framework's latency, computational cost, and real-time performance. \\
While LLMs show promise for engineering support, their direct integration into autonomous decision-making processes for UAVs remains a critical area requiring further research and validation due to safety-critical implications. \\

\appendix
\section{Appendix}

This section presents the appendix, which includes the prompt templates, questions, and corresponding answers related to the models.
\vspace{-.5cm}
\subsection{Prompt template}

We used the following prompt in the RAG process:\\
\textit{
To answer the questions, use only the documents provided.
Formulations are provided for all the questions in the provided documents; however, to calculate the result, you may use your calculation skills.
\\{context}
Question: {question}}
\label{appendix:prompt}
\subsection{Question 1-2: GPT o1 Answer without RAG}
\begin{tcolorbox}[title=Prompt, colback=gray!5, colframe=black!50]
Consider this information:
\begin{itemize}
    \item The constant value $p_0$: $1.5$
    \item The forward speed $v$: $\SI{110}{\kilo\meter\per\hour}$
    \item The blade angular velocity $\Omega$: $\SI{300}{\radian\per\second}$
    \item The rotor radius $R$: $\SI{0.4}{\meter}$
\end{itemize}

Based on this information, compute the Blade power of UAV in forward flight?
\end{tcolorbox}

\vspace{0.5em}

\begin{tcolorbox}[title=o1 Response without RAG, colback=gray!2, colframe=black!30]
A key difficulty with this problem is that, as stated, there is not enough information to arrive at a unique numerical value for the rotor (blade) power in forward flight. In helicopter-type rotor analyses, one typically needs at least:
\begin{itemize}
    \item The vehicle’s weight (or total thrust the rotor must produce).
    \item The air density ($\rho$).
    \item Basic blade-geometry parameters (number of blades, chord, airfoil drag coefficient, etc.).
    \item A suitable rotor-theory formula (e.g., blade-element or momentum theory).
\end{itemize}

Simply giving
\begin{itemize}
    \item $p_0 = 1.5$ (an unspecified “constant”),
    \item forward speed $v = 110\,\text{km/h}$ ($\approx30.6\,\text{m/s}$),
    \item angular velocity $\Omega = 300\,\text{rad/s}$,
    \item rotor radius $R = 0.4\,\text{m}$
\end{itemize}
does not pin down the aerodynamic power.

\section*{Why More Information is Needed}

\begin{enumerate}
    \item In forward flight, total rotor power typically has at least two major components:
    \begin{itemize}
        \item Induced (or “lift-induced”) power, often estimated from momentum theory.
        \item Profile (or “parasite”) power due to drag on the spinning blades themselves, often estimated from blade-element theory.
    \end{itemize}
    \item Both components depend on air density ($\rho$) and on how much thrust the rotor must generate (which in turn depends on the UAV’s weight and flight condition).
    \item Even if one assumes the thrust requirement is known, there are empirical or semi-empirical “constants” in the formulas (such as blade drag coefficients, tip-loss factors, etc.) that must be specified or lumped into a known expression before a numeric answer can come out.
\end{enumerate}

\section*{Typical Example Formulas}

By way of illustration (not a complete solution), in many helicopter texts one finds a “profile-power” formula of the form
\[
P_{\text{profile}} \approx \frac{1}{8} \rho \sigma C_{D_0} \pi R^2 (\Omega R)^3 \times \left[1 + k \left(\frac{v}{\Omega R}\right)^2\right],
\]
where
\begin{itemize}
    \item $\sigma$ is the rotor solidity (depends on blade chord and number of blades),
    \item $C_{D_0}$ is the airfoil’s drag coefficient,
    \item $k$ is an empirical factor ($\approx 4 \text{ to } 6$),
    \item $v$ is forward speed, and
    \item $\Omega R$ is the blade tip speed.
\end{itemize}
Even here, you can see one must know $\rho$, $\sigma$, $C_{D_0}$, and so on.
\section*{Conclusion}

Because the problem as posed does not provide the rotor thrust (or vehicle weight), the air density, or enough blade-geometry/aerodynamic details, there is no way to turn the given numbers into a single “watts” (or kW) value for rotor power. One can only proceed once those other parameters and the specific power formula are specified.
\label{app:answerq1-2}
\end{tcolorbox}

\subsection{Question 1-2: GPT o1 Answer with RAG}
\begin{tcolorbox}[title=Prompt, colback=gray!5, colframe=black!50]
Consider this information:
\begin{itemize}
    \item The constant value $p_0$: $1.5$
    \item The forward speed $v$: $\SI{110}{\kilo\meter\per\hour}$
    \item The blade angular velocity $\Omega$: $\SI{300}{\radian\per\second}$
    \item The rotor radius $R$: $\SI{0.4}{\meter}$
\end{itemize}

Based on this information, compute the Blade power of UAV in forward flight?
\end{tcolorbox}

\vspace{0.5em}

\begin{tcolorbox}[title=o1 Response with RAG,  colback=gray!2, colframe=black!30]
Below is a concise way to obtain the “blade power” term for forward flight from the documents. From (66) in the provided text, the blade-profile (or “blade”) power in forward flight is the first bracketed term, namely
\[
P_{\text{blade}}(V) = P_0 \left( 1 + \frac{3V^2}{\Omega^2 R^2} \right).
\]

Given Data
\begin{itemize}
    \item $P_0 = 1.5$ (constant), Forward speed $V = \SI{110}{\kilo\meter\per\hour}$
    \item Blade angular velocity $\Omega = \SI{300}{\radian\per\second}$, Rotor radius $R = \SI{0.4}{\meter}$
\end{itemize}

Calculation Steps Step 1: Convert V into meters per second \[
V (\si{\meter\per\second}) = 110 \times \frac{1000}{3600} \approx \SI{30.56}{\meter\per\second}.
\]

Step 2: Compute the ratio $V^2 / (\Omega^2 R^2)$
\begin{itemize}
    \item $\Omega^2 R^2 = (\SI{300}{\radian\per\second})^2 \times (\SI{0.4}{\meter})^2 = \num{90000} \times \num{0.16} = \num{14400} \, (\si{\meter\per\second})^2$.
    \item $V^2 \approx (\SI{30.56}{\meter\per\second})^2 \approx \SI{933.9136}{(\meter\per\second)^2}$.
    \item Hence, $\displaystyle \frac{V^2}{\Omega^2 R^2} \approx \frac{933.9136}{14400} \approx 0.064855$.
\end{itemize}
(Using the input's rounded value for $V^2/( \Omega^2 R^2) \approx 0.0649$ for the next step as per the provided text.)

Step 3: Substitute into the blade-power expression: 
Using the approximation $V^2 / (\Omega^2 R^2) \approx 0.0649$:
\[
1 + 3 \times 0.0649 \approx 1 + 0.1947 = 1.1947.
\]
\[
P_{\text{blade}}(V) = 1.5 \times 1.1947 \approx 1.79205.
\]
(Rounding to $1.79$ as per the provided text.)

Result: Thus, the blade power at $V = \SI{110}{\kilo\meter\per\hour}$ is approximately $1.79$ (in the same power units as $P_0$).

\label{app:correct1-2}
\end{tcolorbox}

\subsection{Question 1-1: Deep Seek Answer with RAG}

\begin{tcolorbox}[title=Prompt, colback=gray!5, colframe=black!50]
Consider this information:
\begin{enumerate}
    \item Ground user’s position: (47,23)m
    \item UAV-BS’s position: (65,73,15)m
\end{enumerate}

Based on this information, what speed is elevation angle?
\end{tcolorbox}

\vspace{0.5em}

\begin{tcolorbox}[title=DeepSeek R1 Response,  colback=gray!2, colframe=black!30]
Okay, so I need to figure out the elevation angle based on the given positions of the ground user and the UAV-BS...
Okay, so I need to figure out the elevation angle based on the given positions of the ground user and the UAV-BS. Let me start by understanding what an elevation angle is. From what I remember, the elevation angle in the context of wireless communications or between two points is the angle between the line connecting the two points (in this case, the ground user and the UAV) and the horizontal plane. So it's like looking up from the ground to the UAV, and the angle your line of sight makes with the horizontal.

First, I need to find the distance between these two points in 3D space because the elevation angle depends on their positions. The ground user is at (47,23) meters, which I assume is in a 2D plane, so their z-coordinate is 0. The UAV-BS is at (65,73,15)m, so its x and y coordinates are 65 and 73 respectively, and it's 15 meters above the ground.

To find the elevation angle, I think I can model this as a right triangle where the vertical distance from the ground user to the UAV is one leg (which is 15 meters), and the horizontal distance between them is the other leg. The hypotenuse would then be the straight-line distance between the two points in 3D space.

So, first step: calculate the horizontal distance between the two points. Since both are given in x and y coordinates, I can use the Euclidean distance formula for 2D:
\begin{equation}
\text{Horizontal distance} = \sqrt{(65 - 47)^2 + (73 - 23)^2}
\end{equation}

Let me compute that:

$65 - 47 = 18$ \\
$73 - 23 = 50$

So, horizontal distance =
\begin{equation}
\text{Horizontal distance} = \sqrt{18^2 + 50^2} =\\ \sqrt{324 + 2500} =\\ \sqrt{2824}
\end{equation}

Calculating $\sqrt{2824}$. Let's see: $53^2$ is 2809, so $\sqrt{2824}$ is a bit more than 53. Maybe approximately 53.11 meters.

Now, the vertical distance is given as 15 meters.

So now I have a right triangle with opposite side (vertical) = 15m and adjacent side (horizontal) $\approx 53.11$m.

The elevation angle $\theta$ can be found using the tangent function:
\begin{equation}
\tan(\theta) = \frac{\text{opposite}}{\text{adjacent}} = \frac{15}{53.11}
\end{equation}

Let me compute that: 15 divided by approximately 53.11 is roughly 0.2823.

Now, to find $\theta$, I take the arctangent of 0.2823. Using a calculator, $\arctan(0.2823)$ is about 15.7 degrees.

Wait, let me double-check my calculations because sometimes it's easy to mix up the sides. The elevation angle is indeed calculated using $\tan(\theta) = \text{opposite}/\text{adjacent}$, which in this case is vertical over horizontal distance. So yes, that should be correct.

Alternatively, I could use the sine function since $\sin(\theta) = \frac{\text{opposite}}{\text{hypotenuse}}$. Let me compute the hypotenuse first:
\begin{equation}
\text{Hypotenuse} = \sqrt{\text{horizontal}^2 + \text{vertical}^2} = 
\end{equation}
\\
\begin{equation}
    \sqrt{53.11^2 + 15^2}
\end{equation}
Calculating $53.11$ squared: approx 2824, and $15$ squared is 225, so total under the root is $2824 + 225 = 3049$.
\begin{equation}
\text{Hypotenuse} \approx \sqrt{3049} \approx 55.22 \text{ meters}
\end{equation}

Then
\begin{equation}
\sin(\theta) = \frac{15}{55.22} \approx 0.2717
\end{equation}

Taking arcsin of that gives $\theta \approx 15.75$ degrees, which aligns with my previous calculation using tangent. So both methods give me around 15.7 to 15.8 degrees, so I can be confident it's approximately 15.7 degrees.

I think that's the elevation angle between the ground user and the UAV-BS based on their positions.

The elevation angle between the ground user at (47,23)m and the UAV-BS at (65,73,15)m is calculated as follows:

\textbf{1. Horizontal Distance Calculation:}
\begin{itemize}
    \item $\Delta x = 65 - 47 = 18$m
    \item $\Delta y = 73 - 23 = 50$m
    \item Horizontal distance $= \sqrt{18^2 + 50^2} \approx 53.11$m
\end{itemize}

\textbf{2. Vertical Distance:}
\begin{itemize}
    \item Vertical distance $= 15$m (z-coordinate of UAV-BS)
\end{itemize}

\textbf{3. Elevation Angle Calculation using Tangent:}
\begin{equation}
\tan(\theta) = \frac{\text{opposite}}{\text{adjacent}} = \frac{15}{53.11} \approx 0.2823
\end{equation}
\begin{equation}
\theta \approx \arctan(0.2823) \approx 15.7^\circ
\end{equation}

\textbf{4. Verification using Sine:}
\begin{equation}
\text{Hypotenuse} = \sqrt{53.11^2 + 15^2} \approx 55.22\text{m}
\end{equation}
\begin{equation}
\sin(\theta) = \frac{15}{55.22} \approx 0.2717
\end{equation}
\begin{equation}
\theta \approx \arcsin(0.2717) \approx 15.75^\circ
\end{equation}

\label{app:deepseekq1-1}
\end{tcolorbox}

\end{document}